\documentclass[conference,10pt]{IEEEtran}
\IEEEoverridecommandlockouts
\usepackage{cite}
\usepackage{amsmath,amssymb,amsfonts}
\usepackage{algorithmic}
\usepackage{graphicx}
\usepackage{textcomp}
\def\BibTeX{{\text B\kern-.05em{\sc i\kern-.025em b}\kern-.08em
    T\kern-.1667em\lower.7ex\hbox{E}\kern-.125emX}}
\usepackage{algorithm,algorithmic}
\usepackage{hyperref}
\begin{document}
\title{Adversarial Training for Probabilistic Spiking Neural Networks}
\author{\IEEEauthorblockN{Alireza Bagheri}
\IEEEauthorblockA{\textit{$^\dag$ECE Department} \\
\textit{New Jersey Institute of Technology}\\
Newark, NJ 07102, USA\\
Email: $ab745@njit.edu$}
\and
\IEEEauthorblockN{Osvaldo Simeone$^\dag$}
\IEEEauthorblockA{\textit{Department of Informatics} \\
\textit{King's College London}\\
London, WC2R 2LS, UK\\
Email: $osvaldo.simeone@kcl.ac.uk$}
\and
\IEEEauthorblockN{Bipin Rajendran}
\IEEEauthorblockA{\textit{ECE Department} \\
	\textit{New Jersey Institute of Technology}\\
	Newark, NJ 07102, USA\\
	Email: $bipin@njit.edu$}
}

\maketitle
\begin{abstract}
Classifiers trained using conventional empirical risk minimization or maximum likelihood methods are known to suffer dramatic performance degradations when tested over examples adversarially selected based on knowledge of the classifier's decision rule. Due to the prominence of Artificial Neural Networks (ANNs) as classifiers, their sensitivity to adversarial examples, as well as robust training schemes, have been recently the subject of intense investigation. In this paper, for the first time, the sensitivity of spiking neural networks (SNNs), or third-generation neural networks, to adversarial examples is studied. The study considers rate and time encoding, as well as rate and first-to-spike decoding. Furthermore, a robust training mechanism is proposed that is demonstrated to enhance the performance of SNNs under white-box attacks.
\end{abstract}
%
\section{INTRODUCTION}\label{sec:Intro}
%
The classification accuracy of Artificial Neural Networks (ANNs) trained over large data sets from the problem domain has attained super-human levels for many tasks including image identification \cite{ranjan2018deep}. Nevertheless, the performance of classifiers trained using conventional empirical risk minimization or Maximum Likelihood (ML) is known to decrease dramatically when evaluated over examples adversarially selected based on knowledge of the classifier's decision rule \cite{goodfellow2014explaining}. To mitigate this problem, robust training strategies that are aware of the presence of adversarial perturbations have been shown to improve the accuracy of classifiers, including ANNs, when tested over adversarial examples 
\cite{goodfellow2014explaining, fawzi2017robustness, madry2017towards}.

ANNs are known to be energy-intensive, hindering their implementation on energy-limited processors such as mobile devices. Despite the recent industrial efforts around the production of more energy-efficient chips for ANNs \cite{paugam2012computing}, the gap between the energy efficiency of the human brain and that of ANNs remains significant \cite{Inte_web, smith2017research}. A promising alternative paradigm is offered by Spiking Neural Networks (SNNs), in which synaptic input and neuronal output signals are sparse asynchronous binary spike trains \cite{paugam2012computing}. Unlike ANNs, SNNs are hybrid digital-analog machines that make use of the temporal dimension, not just as a neutral substrate for computing, but as a means to encode and process information \cite{smith2017research}. 

Training methods for SNNs typically assume deterministic non-linear dynamic models for the spiking neurons, and are either motivated by biological plausibility, such as the spike-timing-dependent plasticity (STDP) rule \cite{paugam2012computing, ponulak2010supervised}, or by an attempt to mimic the operation of ANNs and associated learning rules (see, e.g., \cite{sengupta2018going} and references therein). 
Deterministic models are known to be limited in their expressive power, especially as it pertains prior domain knowledge, uncertainty, and definition of generic queries and tasks. Training for probabilistic models of SNNs has recently been investigated in, e.g., \cite{gardner2016supervised, guo2017hierarchical, rezende2011variational, bagheri2017training}  using ML and variational inference principles. 

\begin{figure}[t]
	\centering
	\centerline{\resizebox{0.8\columnwidth}{!}{\includegraphics{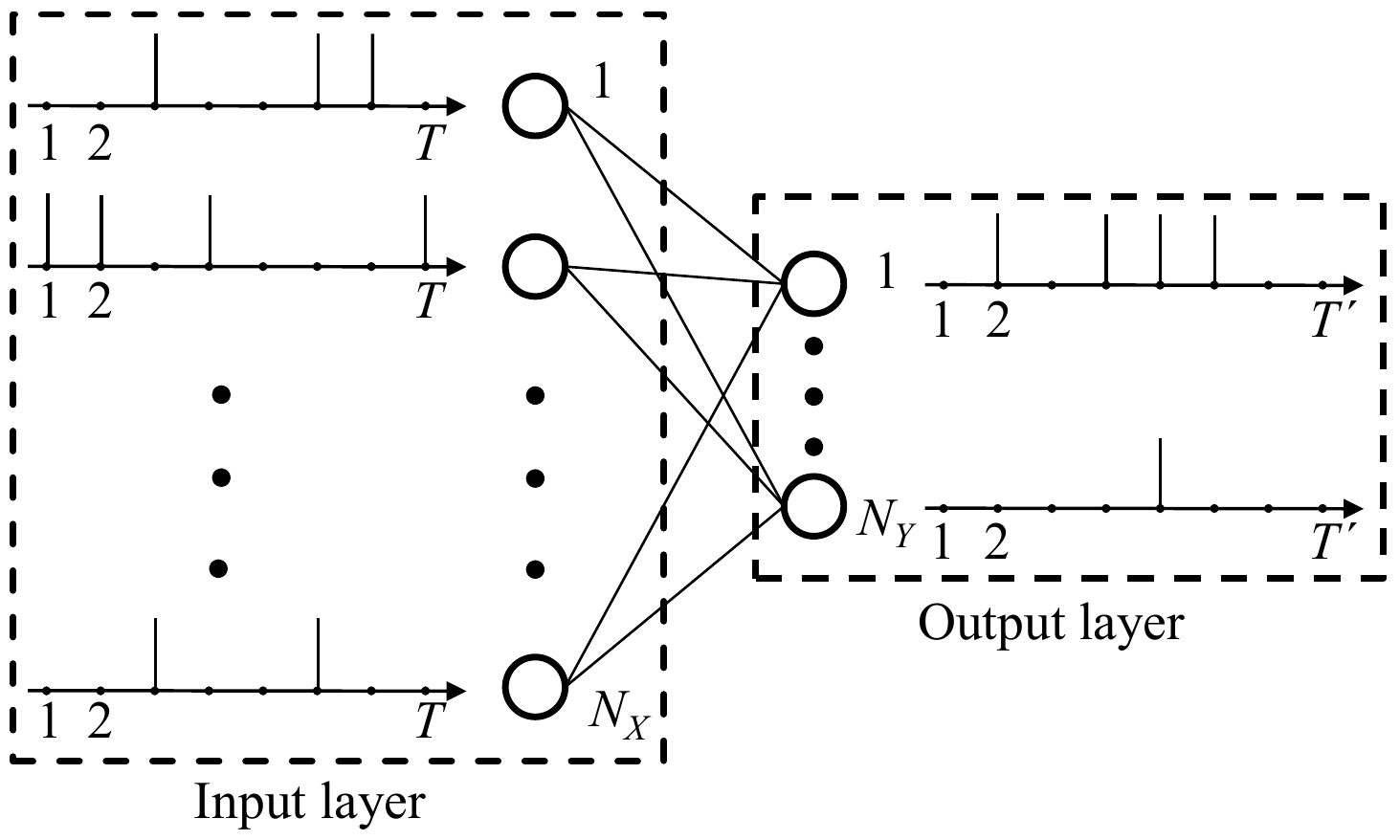}}}	
	\caption{Two-layer SNN for supervised learning.}
	\label{fig:Net_diag}
\end{figure}

In this paper, for the first time, the sensitivity of SNNs trained via ML is studied under white-box adversarial attacks, and a robust training mechanism is proposed that is demonstrated to enhance the performance of SNNs under adversarial examples. Specifically, we focus on a two-layer SNN (see Fig.~\ref{fig:Net_diag}), and consider rate and time encoding, as well as  rate and first-to-spike decoding \cite{bagheri2017training}. Our results illuminate the sensitivity of SNNs to adversarial example under different encoding and decoding schemes, and the effectiveness of robust training methods. 

%
%

The rest of the paper is organized as follows. In Sec. \ref{sec:Architectures}, we describe the architecture of the two-layer SNN with Generalized Linear Model (GLM) neuron, as well as information encoding and decoding mechanisms.
The design of adversarial perturbations is covered in Sec. \ref{sec:Adversarial_Examples}, while a robust training is presented in Sec.  \ref{sec:Robust_training}. Sec. \ref{sec:Sim_sec} presents numerical results, and closing remarks are given in Sec. \ref{sec:conclusion_sec}.

%
\section {SNN-BASED CLASSIFICATION}\label{sec:Architectures}
%
In this section, we introduce the classification task and the SNN architecture under study.

%
{\bf{Network Architecture}}:\label{ssec:Net_Arch}
%
We consider the problem of classification using the two-layer SNN illustrated in Fig.~\ref{fig:Net_diag}. 
The SNN is fully connected and has $N_X$ presynaptic neurons in the input, or sensory layer, and $N_Y$ neurons in the output layer. Each output neuron is associated with a class. In order to feed the SNN, an input example, e.g., a gray scale image, is encoded into a set of $N_X$ discrete-time spike trains, each with $T$ samples. The input spike trains are fed to the $N_Y$ postsynaptic GLM neurons, which output discrete-time spike trains. A decoder then selects the image class on the basis of the spike trains emitted by the output neurons. 

%
{\bf{Information Encoding}}:\label{ssec:Info_enc}
%
We consider two encoding mechanisms.

%
%
$\textit{1})$ \textit{{Rate encoding}}:
%
With the conventional rate encoding method (see, e.g., \cite{stromatias2017event}), each entry of the input signal is converted into a discrete-time spike train by generating an independent and identically distributed (i.i.d.) Bernoulli vector. The probability of generating a $``1"$, i.e., a spike, is proportional to the value of the entry. In the experiments in Sec. \ref{sec:Sim_sec}, we use gray scale images of USPS dataset with pixel intensities normalized between $0$ and $1$ that yield a proportional spike probability between $0$ and $1/2$.

%
$\textit{2})$ \textit{{Time encoding}}:
%
With the time encoding method, each entry of the input signal is converted into a spike train having only one spike, whose timing depends on the entry value. In particular, assuming intensity-to-latency encoding \cite{masquelier2007unsupervised, kheradpisheh2016stdp, stromatias2017event}, the spike timing in the time interval $\left[ {1,T} \right]$ depends linearly on the entry value, such that the maximum value yields a spike at the first time sample $t = 1$, and the minimum value is mapped to a spike in the last time sample $t = T$.

{\bf{GLM Neuron Model}}:\label{ssec:GLM_model}
%
The relationship between the input spike trains from the ${N_X}$ presynaptic neurons and the output spike train of any postsynaptic neuron $i$ follows a Bernoulli GLM with canonical link function (see, e.g., \cite{pillow2008spatio, bagheri2017training}).
To elaborate, we denote as ${x_{j,t}}$ and ${y_{i,t}}$ the binary signal emitted by the $j$-th presynaptic and the $i$-th postsynaptic neurons, respectively, at time $t$. Also, we let ${\bf{x}}_{j,a}^{b} = \left( {{x_{j,a}},...,{x_{j,b}}} \right)$
be the vector of samples from spiking process of the presynaptic neuron $j$ in the time interval $\left[ {a, b} \right]$. Similarly, the vector ${\bf{y}}_{i,a}^{b} = ({y_{i,a}},...,{y_{i,b}})$ contains samples from the spiking process of the neuron $i$ in the interval $\left[ {a, b} \right]$.
The membrane potential of postsynaptic neuron $i$ at time $t$ is given by
\begin{equation} \label{membrane_potential_GLM}
{u_{i,t}} = \sum\limits_{j = 1}^{{N_X}} {{\boldsymbol{\alpha }}_{j,i}^T{\bf{x}}_{j,t - {\tau _y}}^{t - 1}}  + {\boldsymbol{\beta }}_i^T{\bf{y}}_{i,t - {\tau_y^{\prime}}}^{t - 1} + {\gamma _i},
\end{equation}
where ${{\boldsymbol{\alpha }}_{j,i}} \in {\mathbb{R} ^{\tau_y} }$ is a vector that defines the \textit{synaptic kernel} (SK)
applied on the $\left\{ {j,i} \right\}$ synapse between presynaptic neuron  $j$ and postsynaptic neuron $i$; ${{\boldsymbol{\beta }}_i} \in {\mathbb{R}^{\tau _y^{\prime}}}$ is the \textit{feedback kernel} (FK); and ${{\gamma _i}}$ is a bias parameter. Note that ${\tau_y}$ and ${\tau_y^{\prime}}$ denote the lengths of the SK and FK, respectively.
The vector of variable parameters ${{\boldsymbol{\theta }}_i}$ includes the bias ${\gamma _i}$ and the parameters that define the SK and FK filters, which are discussed below.
According to the GLM, the log-probability of the output spike train ${{\bf{y}}_i} = {\left[ {{y_{i,1}},...,{y_{i,T}}} \right]^T}$ conditioned on the input spike trains ${\bf{x}} = \left\{ {{{\bf{x}}_j}} \right\}_{j = 1}^{{N_X}}$ can be written as
\begin{equation} \label{loglikelihood_BGLM}
\log {p_{{{\boldsymbol{\theta }}_i}}}\!\left( {{{\bf{y}}_i}\left| {\bf{x}} \right.} \right) = \sum\limits_{t = 1}^T {\left[ {{y_{i,t}}\log g\left( {{u_{i,t}}} \right) + {{\bar y}_{i,t}}\log \bar g\left( {{u_{i,t}}} \right)} \right]} ,
\end{equation}
where $g\left( \cdot \right)$ is an activation function, such as the sigmoid function $g\left( x \right) = \sigma \left( x \right) = 1/\left( {1 + \exp \left( { - x} \right)} \right)$, and we defined ${{\bar y}_{i,t}} = 1 - {y_{i,t}}$ and $\bar g\left( {{u_{i,t}}} \right) = 1 - g\left( {{u_{i,t}}} \right)$. As per \eqref{loglikelihood_BGLM}, each sample ${{y}_{i,t}}$ is Bernoulli distributed with spiking probability $g\left( {{u_{i,t}}} \right)$.

As in \cite{bagheri2017training}, we adopt the parameterized model of \cite{pillow2008spatio} for the SK and FK filters. Accordingly, we write the SK ${{\boldsymbol{\alpha }}_{j,i}}$ and the FK ${{\boldsymbol{\beta }}_i}$ as
\begin{equation} \label{GLM_Alpha}
{{\boldsymbol{\alpha }}_{j,i}} = \sum\limits_{k = 1}^{{K_{\boldsymbol{\alpha }}}} {{w_{j,i,k}}{{\bf{a}}_k}} = {\bf{A}}{{\bf{w}}_{j,i}} ,
\end{equation}
and
\begin{equation} \label{GLM_Beta}
{{\boldsymbol{\beta }}_{i}} = \sum\limits_{k = 1}^{{K_{\boldsymbol{\beta}} }} {{v_{i,k}}{{\bf{b}}_k}}  = {\bf{B}}{{\bf{v}}_{i}} ,
\end{equation}
respectively, where we have defined the fixed basis matrices ${\bf{A}} = \left[ {{{\bf{a}}_1},...,{{\bf{a}}_{{K_{\boldsymbol{\alpha }}}}}} \right]$ and ${\bf{B}} = \left[ {{{\bf{b}}_1},...,{{\bf{b}}_{{K_{\boldsymbol{\beta }}}}}} \right]$ and the vectors ${{\bf{w}}_{j,i}} = {\left[ {{w_{j,i,1}},...,{w_{j,i,{K_{\boldsymbol{\alpha }}}}}} \right]^T}$ and ${{\bf{v}}_i} = {\left[ {{v_{i,1}},...,{v_{i,{K_{\boldsymbol{\beta }}}}}} \right]^T}$; ${K_{\boldsymbol{\alpha }}}$ and ${K_{\boldsymbol{\beta }}}$ denote the respective number of basis functions; ${{\bf{a}}_k} = {[ {{a_{k,1}},...,{a_{k,{\tau _y}}}} ]^T}$ and ${{\bf{b}}_k} = {[ {{b_{k,1}},...,{b_{k,{\tau _y^{\prime}}}}} ]^T}$ are the basis vectors; and $\left\{ {{w_{j,i,k}}} \right\}$ and $\left\{ {{v_{i,k}}} \right\}$ are the learnable weights for the kernels ${{\boldsymbol{\alpha }}_{j,i}}$ and ${{\boldsymbol{\beta }}_i}$, respectively. 
For the experiments discussed in Sec. \ref{sec:Sim_sec}, we adopt the raised cosine basis functions introduced in \cite[Sec. Methods]{pillow2008spatio}.

%
{\bf{Information Decoding}}:\label{ssec:Info_dec}
%
We also consider two alternative decoding methods, namely rate decoding and first-to-spike decoding. 
$\textit{1})$ \textit{{Rate decoding}}: With rate decoding,  decoding is carried out by selecting the output neuron with the largest number of spikes.
$\textit{2})$ \textit{{First-to-spike decoding}}: With first-to-spike decoding, the class that corresponds to the neuron that spikes first is selected.

%
{\bf{ML training}}:\label{ssec:ML_tr}
%
Conventional ML training is performed differently under rate and first-to-spike decoding methods, as briefly reviewed next.

%
$\textit{1})$ \textit{{Rate decoding}}:
%
With rate decoding, the postsynaptic neuron corresponding to the correct label $c \in \left\{ {1,...,{N_Y}} \right\}$ is assigned a desired output spike train ${\bf{y}}_c$ containing a number of spikes, while an all-zero vector ${\bf{y}}_i$, $i \ne c$, is assigned to the other postsynaptic neurons. Using the ML criterion, one hence maximizes the sum of the log-probabilities \eqref{loglikelihood_BGLM} of the desired output spikes ${\bf{y}}\left( c \right) = \left\{ {{{\bf{y}}_1},...,{{\bf{y}}_{{N_Y}}}} \right\}$ for the given $N_X$ input spike trains ${\bf{x}} = \left\{ {{{\bf{x}}_1},...,{{\bf{x}}_{{N_X}}}} \right\}$.
The log-likelihood function for a given training example $\left( {{\bf{x}}, c} \right)$ can be written as
\begin{equation}\label{Rate_dec_likelihood}
{{L_{\boldsymbol{\theta }}}\left( {{\bf{x}}, c} \right)} = \sum\limits_{i = 1}^{{N_Y}} {\log {p_{{{\boldsymbol{\theta }}_i}}}\!\left( {\left. {{{\bf{y}}_i}} \right|{\bf{x}}} \right)} ,
\end{equation}
where the parameter vector ${\boldsymbol{\theta }} = \left\{ {{\bf{W}},{\bf{V}},{\boldsymbol{\gamma }}} \right\}$ includes the parameters ${\bf{W}} = \left\{ {{{\bf{W}}_i}} \right\}_{i = 1}^{{N_Y}}$, ${\bf{V}} = \left\{ {{{\bf{v}}_i}} \right\}_{i = 1}^{{N_Y}}$ and ${\boldsymbol{\gamma }} = \left\{ {{\gamma _i}} \right\}_{i = 1}^{{N_Y}}$.
The sum in \eqref{Rate_dec_likelihood} is further extended to all examples in the training set. The negative log-likelihood (NLL) $-{L_{\boldsymbol{\theta }}}$ is convex with respect to ${\boldsymbol{\theta }}$ and can be minimized via SGD \cite{bagheri2017training}.

%
$\textit{2})$ \textit{{First-to-spike decoding}}:
%
With first-to-spike decoding, the class that corresponds to the neuron that spikes first is selected. The ML criterion hence maximizes the probability to have the first spike at the output neuron corresponding to the correct label. The logarithm of this probability for a given example $\left( {{\bf{x}}, c} \right)$ can be written as 
%
\begin{equation} \label{FTS_dec_likelihood}
{{L_{\boldsymbol{\theta }}}\left( {{\bf{x}}, c} \right)} = \log \left( {\sum\limits_{t = 1}^T {{p_t}\left( {\boldsymbol{\theta }} \right)} } \right),
\end{equation}
where 
%
\begin{equation} \label{FTS_pt}
{p_t}\left( {\boldsymbol{\theta }} \right) = \prod\limits_{i = 1,i \ne c}^{{N_Y}} {\prod\limits_{t' = 1}^t {\bar g\left( {{u_{i,t'}}} \right)} } g\left( {{u_{c,t}}} \right)\prod\limits_{t' = 1}^{t - 1} {\bar g\left( {{u_{c,t'}}} \right)},
\end{equation}
is the probability of having the first spike at the correct neuron $c$ at time $t$. In \eqref{FTS_pt}, the potential $u_{i,t}$ for all $i$ is obtained from \eqref{membrane_potential_GLM} by setting $y_{i,t}  = 0$ for all $i$ and $t$. The minimization of the log-likelihood function ${L_{\boldsymbol{\theta }}}$ in \eqref{FTS_dec_likelihood}, which is not concave, can be tackled via SGD as proposed in \cite{bagheri2017training}.

%
\section {Designing Adversarial Examples}\label{sec:Adversarial_Examples}
%
In this work, we consider white-box attacks based on full knowledge of the model, i.e., of the parameter vector $\boldsymbol{\theta}$, as well as of the encoding and decoding strategies. Accordingly, given an example $\left( {{\bf{x}},{c}} \right)$, an adversarial spike train ${{\bf{x}}^{{\text{adv}}}}$ is obtained as a perturbed version of the original input ${\bf{x}}$, where the perturbation is selected so as to cause the classifier to be more likely to predict an incorrect label $c' \ne c$, while being sufficiently small.

We consider the following types of perturbations: $(i)$ \textit{Remove attack}: one or more spikes are removed from the input ${\bf{x}}$; $(ii)$ \textit{Add attack}: one or more spikes are added to the input ${\bf{x}}$; and $(iii)$ \textit{Flip attack}: one or more spikes are added or removed. 
The size of the disturbance is measured for all attacks by the number of spikes that are added and/or removed. Mathematically, this can be expressed as the Hamming distance 
\begin{equation} \label{Hamming_dist}
{d_H}\left( {{\bf{x}},{{\bf{x}}^{{\text{adv}}}}} \right) = \sum\limits_{j = 1}^{{N_X}} {\sum\limits_{t = 1}^T {1\left( {{x_{j,t}} \ne x_{j,t}^{{\text{adv}}}} \right)} } ,
\end{equation}
where $1\left( {\cdot} \right)$ is the indicator function, i.e., $1\left( {a} \right) = 1$ if condition $a$ is true and $1\left( {a} \right) = 0$ otherwise.

In order to select the adversarial perturbation of an input ${\bf{x}}$, we consider the maximization of the likelihood of a given incorrect target class $c' \ne c$. According to \cite{kurakin2016adversarial}, an effective way to choose the target class $c'$ is to find the class ${c^{{\text{LL}}}} \ne c$ that is the least likely under the given model $\boldsymbol{\theta}$. Mathematically, for a given training example $\left( {{\bf{x}},c} \right)$, the least likely class is obtained by solving the problem
\begin{equation} \label{c_LL}
\begin{array}{*{20}{l}}
{{c^{{\text{LL}}}} = \mathop {{\text{argmin}}}\limits_{c' \ne c} }&{{L_{\boldsymbol{\theta }}}\left( {{\bf{x}},c'} \right)},
\end{array}
\end{equation}
where the log-likelihood ${{L_{\boldsymbol{\theta }}}\left( {{\bf{x}},c'} \right)}$ is given by \eqref{Rate_dec_likelihood} for rate decoding and \eqref{FTS_dec_likelihood} for first-to-spike decoding.

Then, in order to compute the adversarial perturbation ${\bf{p}}$, we maximize the likelihood of class ${c^{{\text{LL}}}}$ under model $\boldsymbol{\theta}$ by tackling the following optimization problem 
\begin{equation} \label{adv_attack}
\begin{array}{*{20}{l}}
{\mathop {{\text{max}}}\limits_{{\bf{p}} \in \mathcal{C}} }&{{L_{\boldsymbol{\theta }}}\left( {{\bf{x}} + {\bf{p}},{c^{{\text{LL}}}}} \right)}\\
{{\text{s}}{\text{.t}}{\text{.}}}&{{{\left\| {\bf{p}} \right\|}_0} \le \epsilon {N_X}T},
\end{array}
\end{equation}
where ${\left\| \bf{p} \right\|_0}$ denotes the number of non-zero elements of $\bf{p}$. 
In \eqref{adv_attack}, the perturbation $\epsilon > 0$ controls the adversary strength. In particular, the adversary is allowed to add or remove spikes from a fraction $\epsilon$ of the ${N_X}T$ input samples, i.e., $T$ samples for each input neuron. The constraint set $\mathcal{C}$ in problem  \eqref{adv_attack} is given by the set of binary perturbations, i.e., $\mathcal{C} = {\left\{ {0, 1} \right\}^{{N_X}T}}$, for add attacks, since spikes can only be added; $\mathcal{C} = {\left\{ {0, -1} \right\}^{{N_X}T}}$ for remove attacks; and $\mathcal{C} = {\left\{ {0, \pm1} \right\}^{{N_X}T}}$ for flip attacks.

The exact solution of problem \eqref{adv_attack} requires an exhaustive search over all possible perturbations of ${\epsilon {N_X}T}$ samples. In the worst case of flip attacks, the resulting search space is hence exponential in  $N_X$ and $T$. Therefore, here we resort to a greedy search method. As detailed in Algorithm~\ref{Algo_1}, at each of the $\left\lfloor {\epsilon {N_X}T} \right\rfloor$ steps, the method looks for the best spike to add, remove or flip, depending on the attack type. We further reduce complexity by searching only among the first $T_A \leq T$ samples across all input neurons. As a results, the complexity of each step of Algorithm~\ref{Algo_1} is at most ${N_X}{T_A}$.

%
\begin{algorithm}[t]
	\caption{Greedy Design $\left( {{\boldsymbol{\theta }}, {T_A},\epsilon } \right)$}
	\label{Algo_1}
	\begin{algorithmic}[1]
		\renewcommand{\algorithmicrequire}{\textbf{Input:}}\REQUIRE ${{\bf{x}}}$, ${\boldsymbol{\theta }}$,  ${T_A}$, $\epsilon$
		\STATE Compute ${c^{{\text{LL}}}}$ from \eqref{c_LL}
		\STATE Initialize: ${{\bf{x}}^{{\text{adv}}}}\left( 0 \right) \leftarrow  {\bf{x}}$
		\FOR {$i = 1$ \TO $\left\lfloor {\epsilon {N_X}T} \right\rfloor$}
		{
			\STATE ${{\bf{x}}^{{\text{adv}}}}\left( i \right) \leftarrow {{\bf{x}}^{{\text{adv}}}}\left( {i - 1} \right) + {\bf{p}}$, where ${\bf{p}}$ is obtained by solving problem \eqref{adv_attack} with ${{\bf{x}}^{{\rm{adv}}}}\left( {i - 1} \right)$ in lieu of ${\bf{x}}$ and $p_{j, t} = 0$ for all $t > T_A$.
		}\ENDFOR
		\renewcommand{\algorithmicrequire}{\textbf{Output:}}\REQUIRE ${\bf{x}}^{{\text{adv}}}$
	\end{algorithmic}
\end{algorithm}
%

%
\begin{algorithm}[t]
	\caption{Adversarial Training $\left( {{T_A},{\epsilon_A}} \right)$}
	\label{Algo_2}
	\begin{algorithmic}[1]
		\renewcommand{\algorithmicrequire}{\textbf{Input:}}\REQUIRE Training set, basis functions ${\bf{A}}$ and ${\bf{B}}$, learning rate ${\eta}$, ${T_A}$, and  ${\epsilon _A}$
		\renewcommand{\algorithmicrequire}{\textbf{Initialize}:}
		\REQUIRE $\boldsymbol{\theta}$
		\FOR {\text{each iteration}}
		{
			\STATE Choose example $\left( {{\bf{x}},{c}} \right)$ from the training set
			\STATE Compute ${\bf{x}}^{{\text{adv}}}$ and ${c^{{\text{LL}}}}$ from Algorithm~\ref{Algo_1} with input $\boldsymbol{\theta}$, ${T_A}$ and  ${\epsilon _A}$
			\STATE Update $\boldsymbol{\theta}$: $\boldsymbol{\theta} \leftarrow \boldsymbol{\theta} +  \eta {\nabla _{\boldsymbol{\theta }}}{L_{\boldsymbol{\theta }}}\left( {{\bf{x}}^{{\rm{adv}}},{c}} \right)$
		}\ENDFOR
		\renewcommand{\algorithmicrequire}{\textbf{Output:}}\REQUIRE $\boldsymbol{\theta}$	
	\end{algorithmic}
\end{algorithm}
%

%
\section {Robust Training}\label{sec:Robust_training}
%

In order to increase the robustness of the trained SNN to adversarial examples, in this section, we propose a robust training procedure. Accordingly, in a manner similar to \cite{madry2017towards}, during the SGD-based training phase, each training example $\left( {{\bf{x}},{c}} \right)$ is substituted with the adversarial example ${\bf{x}}^{{\text{adv}}}$ obtained from Algorithm~\ref{Algo_1} for the current iterate $\boldsymbol{\theta}$. The training algorithm is detailed in Algorithm~\ref{Algo_2}.
Note that, the robust training algorithm is parameterized by $T_A$ and $\epsilon_A$, which determine the parameters of the assumed adversary during training.

%
\section{Numerical Results}\label{sec:Sim_sec}
%

In this section, we numerically study the performance of the described probabilistic SNN under the adversarial attacks. We use the standard USPS dataset as the input data. As a result, we have $N_X = 256$, with one input neuron per pixel of the $16 \times 16$ images. 
Unless stated otherwise, we focus solely in the classes $\{1, 5, 7, 9 \}$ and we set $T = K = 16$. We assume the worst-case $T_A = T$ for the adversary during the test phase.
For rate decoding, we use a desired spike train with one spike after every three zeros. SGD is applied for 200 training epochs and early stopping is used for all schemes. Holdout validation with 20$\%$ of training samples is applied to select between $10^{-3}$ and $10^{-4}$ for the constant learning rate $\eta$. The model parameters $\boldsymbol{\theta}$ are randomly initialized with uniform distribution between -1 and 1. 

%
\begin{figure}[t]
	\centering
	\centerline{\resizebox{1\columnwidth}{!}{\includegraphics{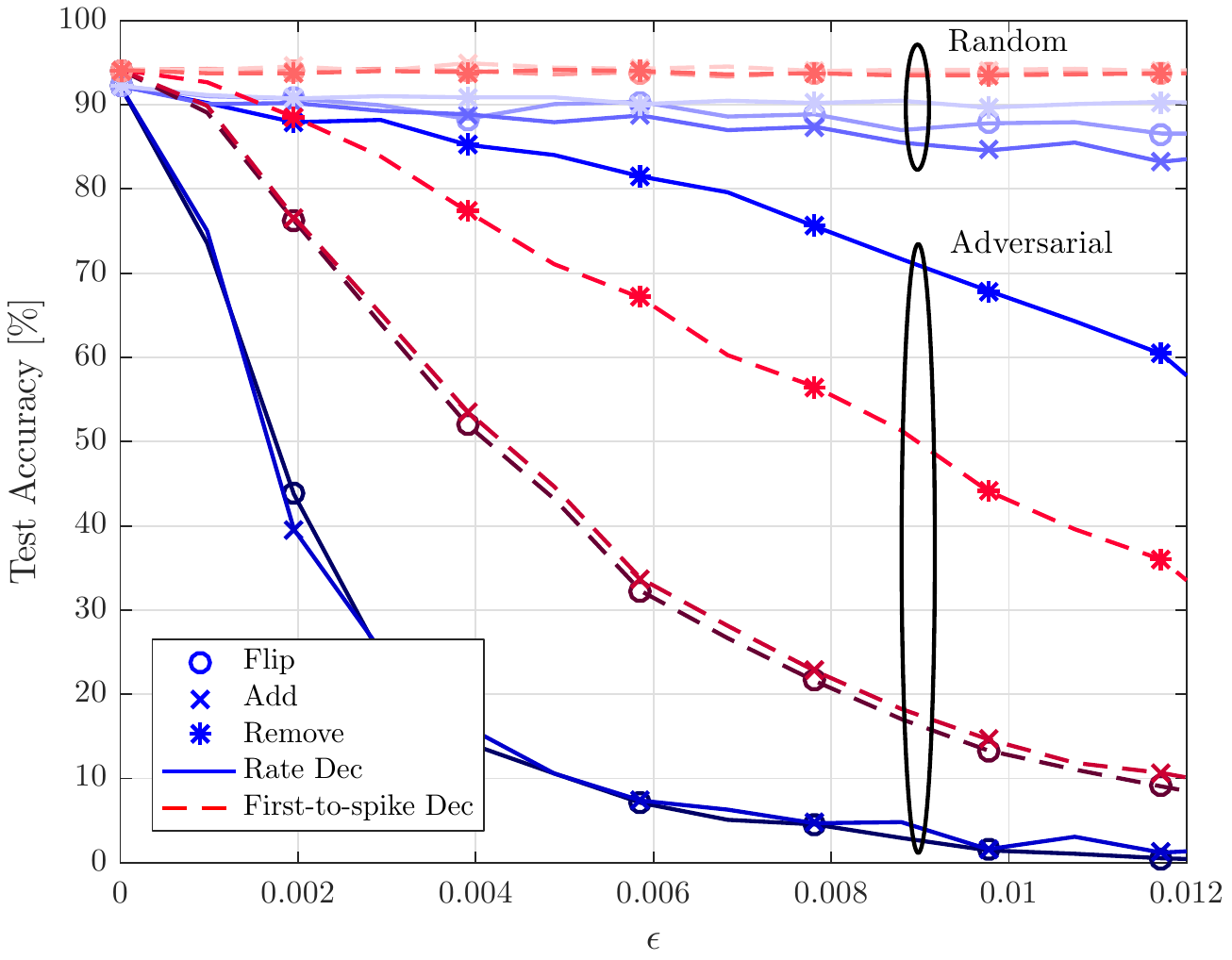}}}	
	\caption{{\small{Test accuracy for ML training under adversarial and random changes versus $\epsilon$ with rate encoding for both rate and first-to-spike decoding rules $\left( T = K = 16 \right)$.}}}
	\label{fig:Dec_vs_eps_RateEnc_ConvFTS_Dec_T_16}
\end{figure}
%

%
\begin{figure}[t]
	\centering
	\centerline{\resizebox{1\columnwidth}{!}{\includegraphics{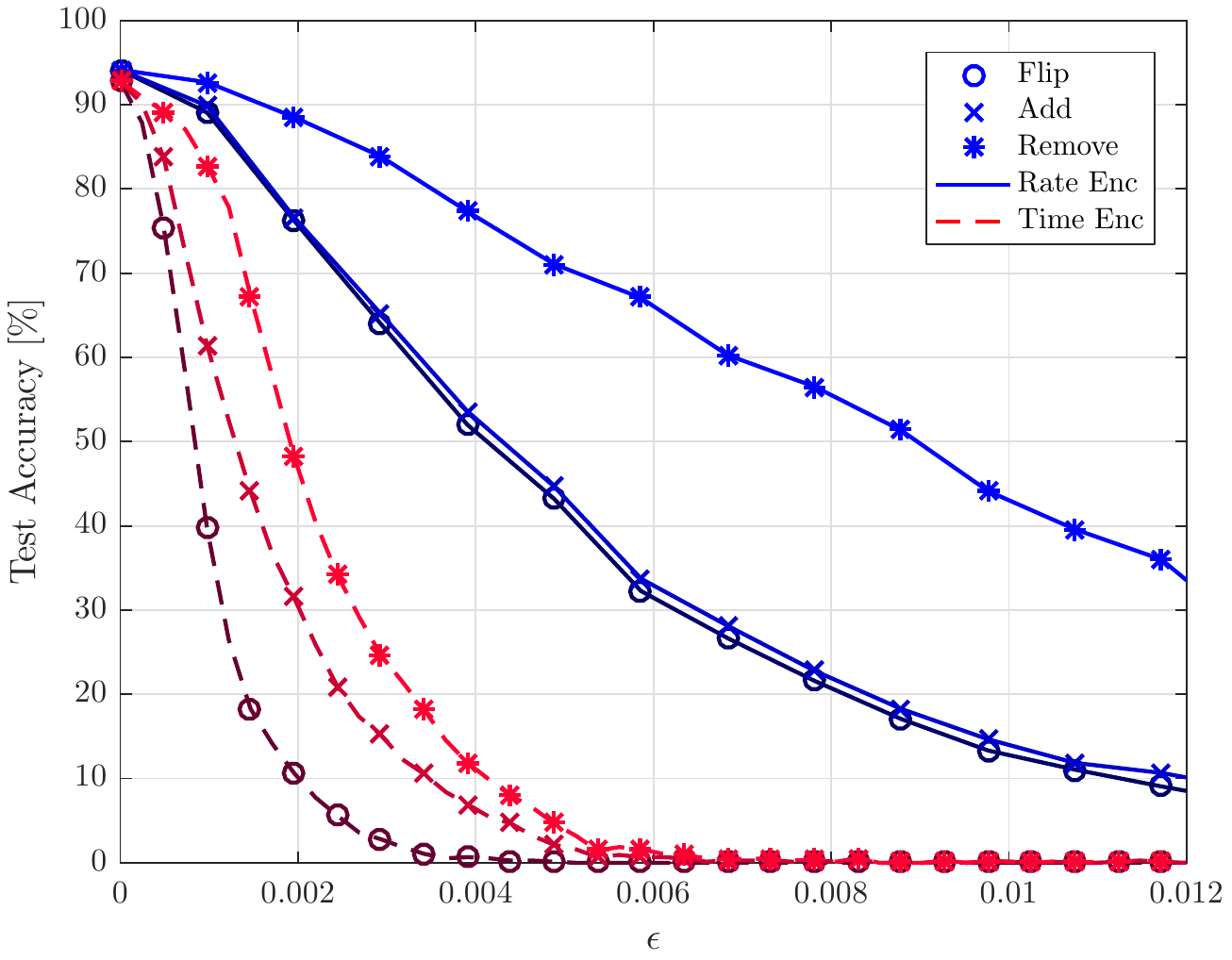}}}	
	\caption{{\small{Test accuracy for ML training under adversarial attacks versus $\epsilon$ with both rate and time encoding rules for first-to-spike decoding $\left( T = K = 16 \right)$.}}}
	\label{fig:Dec_vs_eps_RTEnc_FTS_Dec_T_16}
\end{figure}
%

We first evaluate the sensitivity of different encoding and decoding schemes to adversarial examples obtained as explained in Sec.~\ref{sec:Adversarial_Examples}. For reference, we consider also perturbations obtained by randomly and uniformly adding, removing and flipping spikes.
Fig.~\ref{fig:Dec_vs_eps_RateEnc_ConvFTS_Dec_T_16} illustrates the test accuracy under adversarial and random perturbations when performing standard ML training. The accuracy is plotted versus the adversary's  power $\epsilon$ assuming rate encoding and both rate and first-to-spike decoding rules. The results highlight the notable difference in performance degradation caused by random perturbations and adversarial attacks. In particular, adversarial changes can cause a significant drop in classification accuracy even with small values of $\epsilon$, particularly when the most powerful flip attacks are used. 

First-to-spike decoding is seen to be more resistant to add and flip attacks, while it is more vulnerable than rate decoding to remove spike attacks. The resilience of first-to-spike decoding can be interpreted as a consequence of the fact that the log-likelihood \eqref{FTS_dec_likelihood}, unlike \eqref{Rate_dec_likelihood} for rate decoding, associates multiple outputs to the correct class, namely all of those with the correct neuron spiking first. Nevertheless, removing properly selected spikes can be more deleterious to first-to-spike decoding as it may prevent spiking by the correct neuron.

The comparison between rate and time encoding in terms of sensitivity to adversarial examples is considered in Fig.~\ref{fig:Dec_vs_eps_RTEnc_FTS_Dec_T_16} under the assumption of first-to-spike decoding. Time encoding is seen to be significantly less resilient than rate encoding. This is due to the fact that time encoding, in the form considered here of intensity-to-latency encoding, which associated a single spike per input neuron \cite{stromatias2017event}, can be easily made ineffective by removing selected spikes.

\begin{figure}[t]
	\centering
	\centerline{\resizebox{1\columnwidth}{!}{\includegraphics{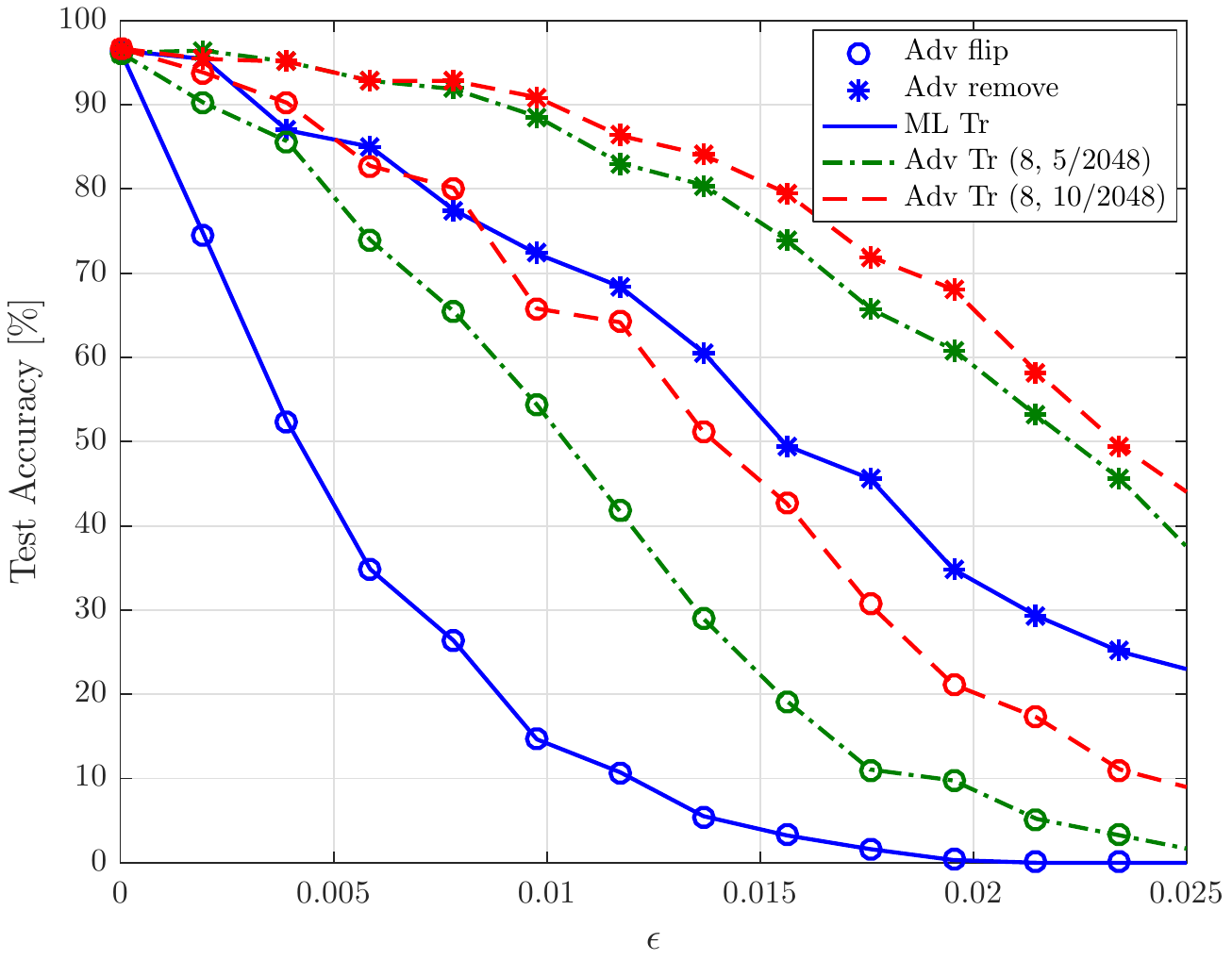}}}	
	\caption{{\small{Test accuracy under adversarial attacks versus $\epsilon$ with rate encoding and rate decoding with ML and adversarial training $\left( T = K = 8 \right)$.}}}
	\label{fig:Dec_vs_eps_Adv_Tr_eps_5_10_RateEnc}
\end{figure}

We then evaluate the impact of robust adversarial training as compared to standard ML. To this end, in Fig.~\ref{fig:Dec_vs_eps_Adv_Tr_eps_5_10_RateEnc}, we plot the test accuracy for the case of flip and remove attacks for both ML and adversarial training when $T = K = 8$. Here we also focus solely on the two classes $\{5, 7\}$. We recall that the adversarial training scheme is parametrized by the time support $T_A$ of the attacks considered during training, here $T_A = 8$, and by its power $\epsilon_A$, here $\epsilon_A = 5/2048$ and $\epsilon_A = 10/2048$. It is observed that robust training can significantly improve the robustness of the SNN classifier, even when $\epsilon_A$ is not equal to the value $\epsilon$ used by the attacker during the test phase. Furthermore, increasing $\epsilon_A$ enhances the robustness of the trained SNN at the cost of a higher computational complexity. 
For instance, for an attacker in the test phase with $\epsilon = 10/2048$, i.e., with 10 bit flips, conventional ML achieves an accuracy of $45\%$, while adversarial training with $\epsilon_A = 10/2048$ (i.e., 10 bit flips) achieves an accuracy of $87\%$. 
Finally, the results show that the classifier remains resilient against other type of attacks, despite being trained assuming the flip attack.

%
%

\begin{figure}[t]
	\centering
	\centerline{\resizebox{1\columnwidth}{!}{\includegraphics{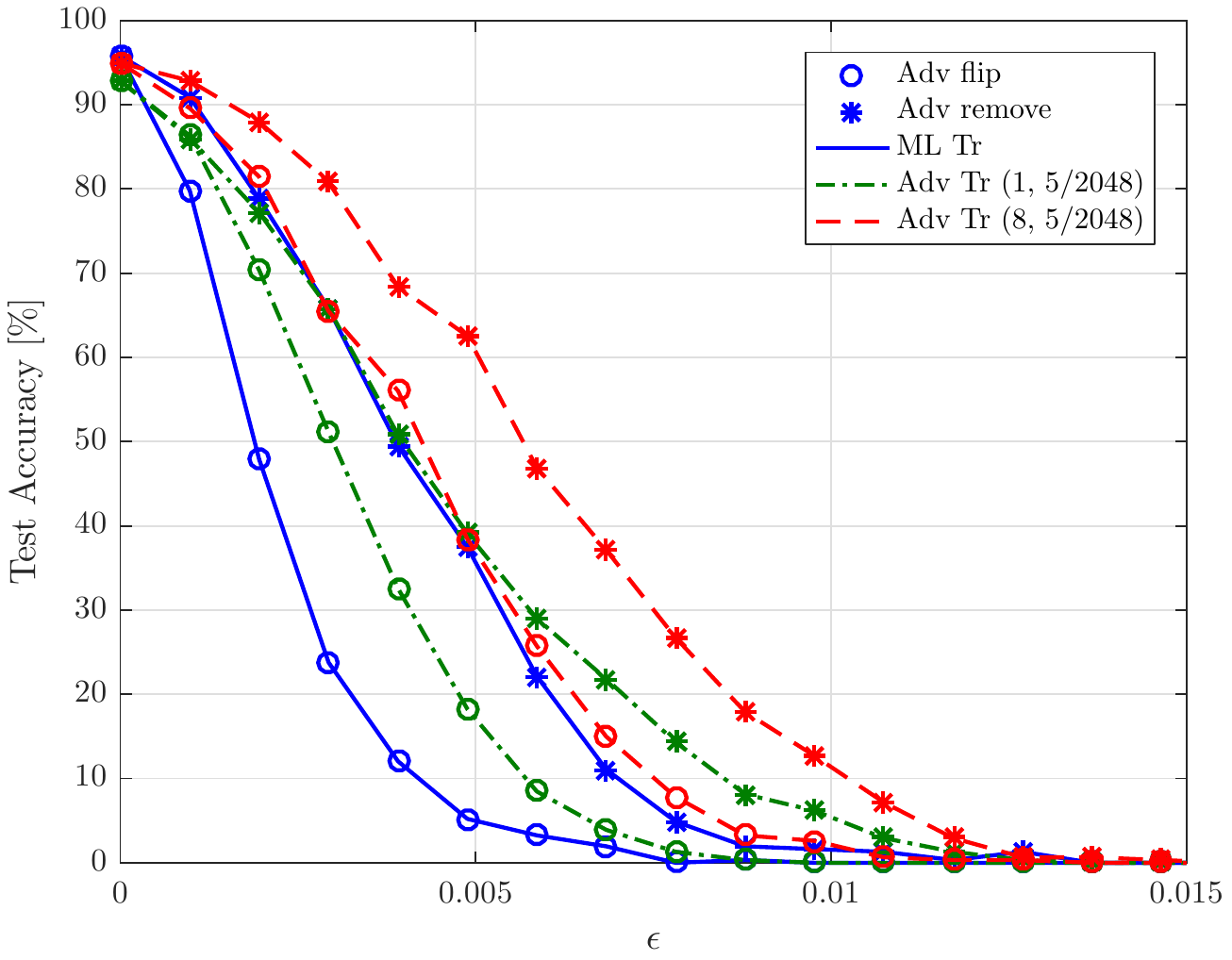}}}	
	\caption{{\small{Test accuracy under adversarial attacks versus $\epsilon$ with time encoding and rate decoding with ML and adversarial training $\left( T = K = 8 \right)$.}}}	\label{fig:Dec_vs_eps_Adv_Tr_t_1_T_TimeEnc_SL}
\end{figure}

Finally, under the same conditions as in Fig.~\ref{fig:Dec_vs_eps_Adv_Tr_t_1_T_TimeEnc_SL}, we study the effect of limiting the power of the adversary assumed during training by considering $T_A = 1$ and $T_A = 8$ with the same $\epsilon_A = 5/2048$. We assume time encoding and rate decoding. It is observed that robust training can still improve the robustness of the SNN classifier, even when $T_A \ll T$ during training. 
For instance, for an attacker in the test phase with $\epsilon = 5/2048$, i.e., 5 bit flips, conventional ML achieves an accuracy of $34.2\%$, while adversarial training with $\epsilon_A = 5/2048$ and $T_A = 1$ and $8$ achieves accuracy levels of $60.3\%$ and $77.5\%$, respectively.

%
\section{Conclusions}\label{sec:conclusion_sec}
%
In this paper, we have studied for the first time the sensitivity of a probabilistic two-layer SNN under adversarial perturbations. 
We considered rate and time encoding, as well as rate and first-to-spike decoding. 
We have proposed mechanisms to build adversarial examples, as well as a robust training method that increases the resilience of the SNN. 
Additional work is needed in order to generalize the results to multi-layer networks.

\section{ACKNOWLEDGMENT}
This work was supported by the U.S. NSF under grant ECCS $\#$1710009. O. Simeone has also received funding from the European Research Council (ERC) under the European Union’s Horizon 2020 research and innovation program (grant agreement $\#$725731).

%
%
\bibliographystyle{IEEEtran}
\bibliography{ieeetran}

\begin{thebibliography}{10}
\providecommand{\url}[1]{#1}
\csname url@samestyle\endcsname
\providecommand{\newblock}{\relax}
\providecommand{\bibinfo}[2]{#2}
\providecommand{\BIBentrySTDinterwordspacing}{\spaceskip=0pt\relax}
\providecommand{\BIBentryALTinterwordstretchfactor}{4}
\providecommand{\BIBentryALTinterwordspacing}{\spaceskip=\fontdimen2\font plus
\BIBentryALTinterwordstretchfactor\fontdimen3\font minus
  \fontdimen4\font\relax}
\providecommand{\BIBforeignlanguage}[2]{{%
\expandafter\ifx\csname l@#1\endcsname\relax
\typeout{** WARNING: IEEEtran.bst: No hyphenation pattern has been}%
\typeout{** loaded for the language `#1'. Using the pattern for}%
\typeout{** the default language instead.}%
\else
\language=\csname l@#1\endcsname
\fi
#2}}
\providecommand{\BIBdecl}{\relax}
\BIBdecl

\bibitem{ranjan2018deep}
R.~Ranjan, S.~Sankaranarayanan, A.~Bansal, N.~Bodla, J.-C. Chen, V.~M. Patel,
  C.~D. Castillo, and R.~Chellappa, ``Deep learning for understanding faces:
  Machines may be just as good, or better, than humans,'' \emph{IEEE Signal
  Process. Mag.}, vol.~35, no.~1, pp. 66--83, 2018.

\bibitem{goodfellow2014explaining}
I.~J. Goodfellow, J.~Shlens, and C.~Szegedy, ``Explaining and harnessing
  adversarial examples,'' in \emph{Int. Conf. on Learn. Repr. (ICLR)}, 2015.

\bibitem{fawzi2017robustness}
A.~Fawzi, S.-M. Moosavi-Dezfooli, and P.~Frossard, ``The robustness of deep
  networks: A geometrical perspective,'' \emph{IEEE Signal Process. Mag.},
  vol.~34, no.~6, pp. 50--62, 2017.

\bibitem{madry2017towards}
A.~Madry, A.~Makelov, L.~Schmidt, D.~Tsipras, and A.~Vladu, ``Towards deep
  learning models resistant to adversarial attacks,'' \emph{arXiv preprint
  arXiv:1706.06083}, 2017.

\bibitem{paugam2012computing}
H.~Paugam-Moisy and S.~Bohte, ``Computing with spiking neuron networks,''
  \emph{Handbook of natural computing}, pp. 335--376, 2012.

\bibitem{Inte_web}
J.~Vincent, ``Intel investigates chips designed like your brain to turn the
  {AI} tide,''
  \textit{https://www.theverge.com/2017/9/26/16365390/intel-investigates-chips-designed-like-your-brain-to-turn-the-ai-tide},
  {A}ccessed: Sept. 26, 2017.

\bibitem{smith2017research}
J.~E. Smith, ``Research agenda: Spacetime computation and the neocortex,''
  \emph{IEEE Micro}, vol.~37, no.~1, pp. 8--14, 2017.

\bibitem{ponulak2010supervised}
F.~Ponulak and A.~Kasi{\'n}ski, ``Supervised learning in spiking neural
  networks with {ReSuMe}: sequence learning, classification, and spike
  shifting,'' \emph{Neural Comput}, vol.~22, no.~2, pp. 467--510, 2010.

\bibitem{sengupta2018going}
A.~Sengupta, Y.~Ye, R.~Wang, C.~Liu, and K.~Roy, ``Going deeper in spiking
  neural networks: {VGG} and residual architectures,'' \emph{arXiv preprint
  arXiv:1802.02627}, 2018.

\bibitem{gardner2016supervised}
B.~Gardner and A.~Gr{\"u}ning, ``Supervised learning in spiking neural networks
  for precise temporal encoding,'' \emph{PloS one}, vol.~11, no.~8, pp. 1--28,
  2016.

\bibitem{guo2017hierarchical}
S.~Guo, Z.~Yu, F.~Deng, X.~Hu, and F.~Chen, ``Hierarchical bayesian inference
  and learning in spiking neural networks,'' \emph{IEEE Trans. Cybern.},
  vol.~PP, no.~99, pp. 1--13, 2017.

\bibitem{rezende2011variational}
D.~J. Rezende, D.~Wierstra, and W.~Gerstner, ``Variational learning for
  recurrent spiking networks,'' \emph{Adv Neural Inf Process Syst}, pp.
  136--144, 2011.

\bibitem{bagheri2017training}
A.~Bagheri, O.~Simeone, and B.~Rajendran, ``Training probabilistic spiking
  neural networks with first-to-spike decoding,'' \emph{arXiv preprint
  arXiv:1710.10704}, 2017.

\bibitem{stromatias2017event}
E.~Stromatias, M.~Soto, T.~Serrano-Gotarredona, and B.~Linares-Barranco, ``An
  event-driven classifier for spiking neural networks fed with synthetic or
  dynamic vision sensor data,'' \emph{Front Neurosci}, vol.~11, pp. 1--17,
  2017.

\bibitem{masquelier2007unsupervised}
T.~Masquelier and S.~J. Thorpe, ``Unsupervised learning of visual features
  through spike timing dependent plasticity,'' \emph{PLoS Comput. Biol.},
  vol.~3, no.~2, pp. 247--257, 2007.

\bibitem{kheradpisheh2016stdp}
S.~R. Kheradpisheh, M.~Ganjtabesh, S.~J. Thorpe, and T.~Masquelier,
  ``{STDP}-based spiking deep neural networks for object recognition,''
  \emph{arXiv preprint arXiv:1611.01421}, 2016.

\bibitem{pillow2008spatio}
J.~W. Pillow, J.~Shlens, L.~Paninski, A.~Sher, A.~M. Litke, E.~Chichilnisky,
  and E.~P. Simoncelli, ``Spatio-temporal correlations and visual signaling in
  a complete neuronal population,'' \emph{Nature}, vol. 454, no. 7207, p. 995,
  2008.

\bibitem{kurakin2016adversarial}
A.~Kurakin, I.~Goodfellow, and S.~Bengio, ``Adversarial examples in the
  physical world,'' \emph{arXiv preprint arXiv:1607.02533}, 2016.

\end{thebibliography}
\end{document}